\documentclass[conference]{IEEEtran}

\usepackage{graphicx}
\usepackage{epstopdf}
\usepackage{amsfonts}
\usepackage{amsmath}
\usepackage{amssymb}
\usepackage[table]{xcolor}
\usepackage[colorinlistoftodos]{todonotes}
\usepackage{flushend}

\newtheorem{definition}{Definition}

\usepackage{cite}
\ifCLASSINFOpdf
\else
\fi
\usepackage{url}
\begin{document}
% \title{Collegare Pallini con gli Algoritmi Evolutivi}

%\title{Accurate and Efficient Inference of\\Cancer Progression Models}
\title{Parallel Implementation of Efficient Search Schemes for the Inference of Cancer Progression Models}

% % author names and affiliations
% % use a multiple column layout for up to three different
% % affiliations
% \author{\IEEEauthorblockN{Daniele Ramazzotti}
% \IEEEauthorblockA{Department of Informatics,\\Systems and Communication\\
% University of Milano-Bicocca\\
% Milano, Italy\\
% Email: ramazzotti@disco.unimib.it}
% \and
% \IEEEauthorblockN{Marco S. Nobile}
% \IEEEauthorblockA{Department of Informatics,\\Systems and Communication\\
% University of Milano-Bicocca\\
% Milano, Italy\\
% Email: nobile@disco.unimib.it}
% \and
% \IEEEauthorblockN{P. Cazzaniga}
% \IEEEauthorblockA{Department of Human Sciences\\
% University of Bergamo\\
% Bergamo, Italy\\
% Email: paolo.cazzaniga@unibg.it}
% \and
% \IEEEauthorblockN{G. Mauri}
% \IEEEauthorblockA{Department of Informatics,\\Systems and Communication\\
% University of Milano-Bicocca\\
% Milano, Italy\\
% Email: mauri@disco.unimib.it}
% \and
% \IEEEauthorblockN{M. Antoniotti}
% \IEEEauthorblockA{Department of Informatics,\\Systems and Communication\\
% University of Milano-Bicocca\\
% Milano, Italy\\
% Email: antoniotti@disco.unimib.it}
% }


\author{\IEEEauthorblockN{Daniele Ramazzotti\IEEEauthorrefmark{1}\IEEEauthorrefmark{2},
Marco S. Nobile\IEEEauthorrefmark{1}\IEEEauthorrefmark{3},
Paolo Cazzaniga\IEEEauthorrefmark{3}\IEEEauthorrefmark{4}, 
Giancarlo Mauri\IEEEauthorrefmark{1}\IEEEauthorrefmark{3} and
Marco Antoniotti\IEEEauthorrefmark{1}}
\IEEEauthorblockA{\IEEEauthorrefmark{1}Department of Informatics, Systems and Communication, University of Milano-Bicocca, Milano, Italy\\ Email: daniele.ramazzotti/nobile/mauri/antoniotti@disco.unimib.it}
\IEEEauthorblockA{\IEEEauthorrefmark{2}Department of Pathology, Stanford University, California 94305, USA}
\IEEEauthorblockA{\IEEEauthorrefmark{3}SYSBIO.IT Centre of Systems Biology, Milano, Italy}
\IEEEauthorblockA{\IEEEauthorrefmark{4}Department of Human and Social Sciences, University of Bergamo, Bergamo, Italy\\ Email: paolo.cazzaniga@unibg.it}}


\maketitle

\begin{abstract}

The emergence and development of cancer is a consequence of the accumulation over time of genomic mutations involving a specific set of genes, which provides the cancer clones with a functional selective advantage. 
In this work, we model the order of accumulation of such mutations during the progression, which eventually  leads to the disease, by means of probabilistic graphic models, i.e., Bayesian Networks (BNs). 
We investigate how to perform the task of learning the structure of such BNs, according to  experimental evidence, adopting a global optimization meta-heuristics.
In particular, in this work we rely on Genetic Algorithms,
and to strongly reduce the execution time of the inference---which can also involve multiple repetitions to collect statistically significant assessments of the data---we distribute the calculations using both multi-threading and a multi-node architecture. 
The results show that our approach is characterized by good accuracy and specificity; we also demonstrate its feasibility, thanks to  a $84 \times$ reduction of the overall execution time with respect to a traditional sequential implementation. 

\end{abstract}


% \fi
\IEEEpeerreviewmaketitle

\section{Introduction}

Cancer development is driven by the subsequent accumulation of genomic mutations over a set of \emph{driver} genes, which confer a functional selective advantage to the cancer clones, leading to the emergence and further development of the disease. 
Indeed, during clonal expansions, tumor cells compete for space and resources and only the fittest clones are capable of outgrowing the competing cells \cite{weinberg2000hallmarks,hanahan2011hallmarks}. 
Here, we aim at modeling such systems in terms of dynamic processes of monotonic accumulations of driver alterations over time. 

Bayesian Networks (BNs) can be exploited to describe the dynamics of biological phenomena characterized by the monotonic accumulation of events, e.g., gene mutations in the context of cancer progression \cite{ramazzotti2016modeling}.
Following this methodology, a temporal ordering among events is implied, so that the occurrence of an early event positively correlates with the subsequent occurrence of its successors.  Such an approach has recently been proved to be effective in \cite{caravagna2015algorithmic}, where progression models are inferred on a cohort of patients derived from two subtypes of colorectal tumors.
% These characteristics make us define a set of properties that we want our reconstructed Bayesian network to be constrained for and lead us to adopt a subset of the general Bayesian networks which we dub Suppes-Bayes Causal networks \cite{}. 

In this work, we first focus on the problem of inferring a BN---which is a well-known NP-hard problem \cite{chickering2004large}---along with the characterization of its complexity and pitfalls. 
Then, we describe possible heuristics to perform the inference of the network and present an efficient parallel implementation of this procedure based on Genetic Algorithms (GAs). 
Finally, we present the results obtained by the application of this approach to synthetic data generated by realistic statistical models, pointing out the satisfactory performance in terms of structural distance from the generative synthetic ground truth and the improvement in execution time achieved thanks to the proposed parallel implementation. 

The paper is structured as follows.
In Section \ref{sec:methods} we describe the problem of the BN inference for the specific problem of cancer progression. 
We then describe the strategy adopted to tackle this problem (GAs), as well as the parallel architecture used to accelerate the inference process.
In Section \ref{sec:results} we present the results obtained from the inference process and the computational speed-up achieved.
We conclude with some remarks and future developments of this work.

\section{Methods}\label{sec:methods}

\subsection{Bayesian Networks and Cancer Progression}
Bayesian Networks are probabilistic graphical models, encoded as Directed Acyclic Graphs (DAGs), which describe the conditional dependence relations among random variables.
Formally, a BN is defined as a DAG $G=(V,E)$, where $V$ is the set of nodes representing any considered random variable, and  $E$ is the set of arcs describing the conditional dependencies among nodes \cite{Koller2009}. 

\begin{figure}[!ht]
  \centering
  \includegraphics[clip,trim=100 100 100 80,width=0.48\textwidth]{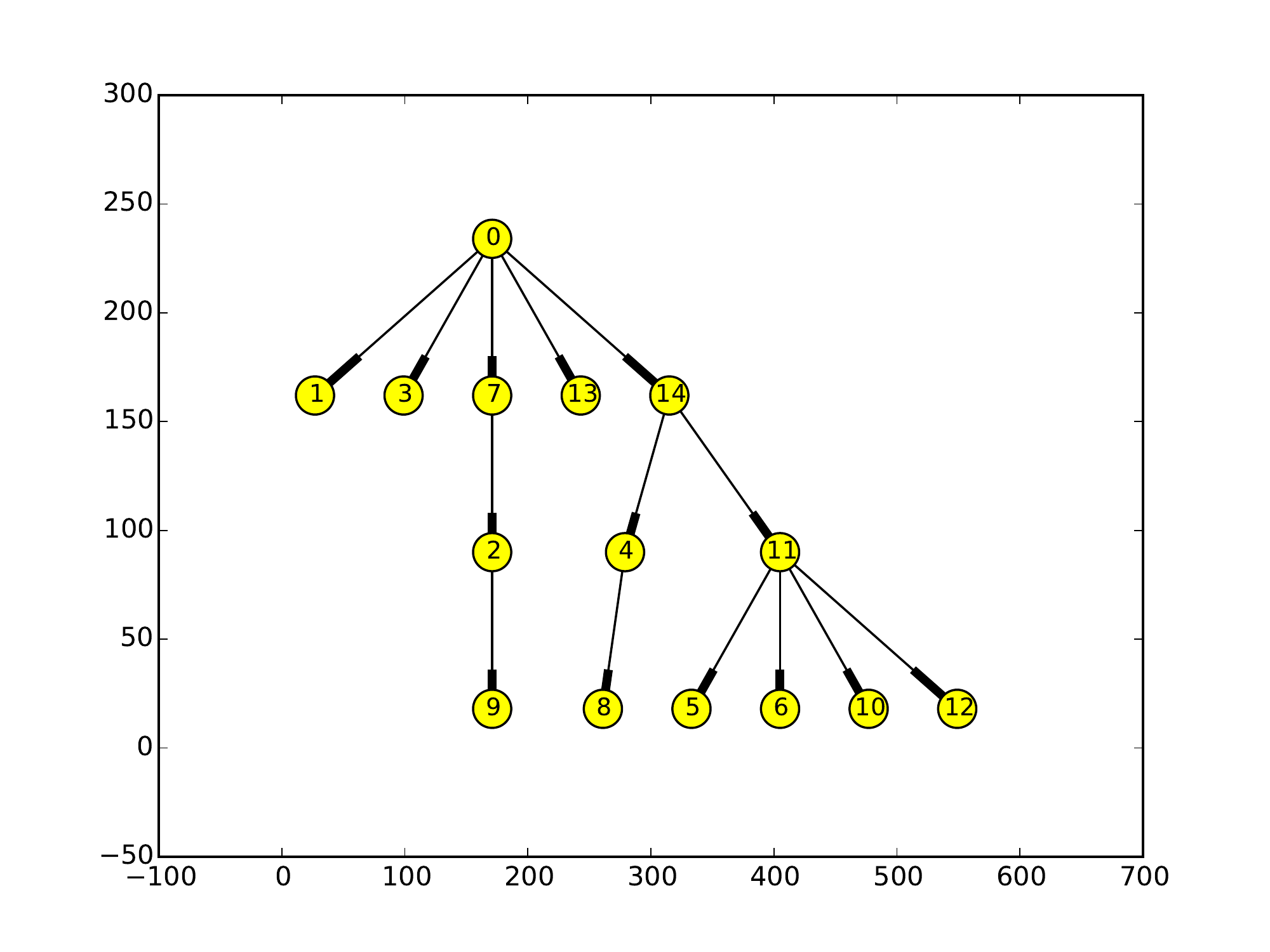}
  \caption{Example of a cancer progression model.
  The nodes in the graph represent gene mutations, while the arcs describe the conditional dependency between the mutations.}
  \label{fig:bn}
\end{figure}

Recently, several statistical methods have been proposed to exploit such Bayesian graphical models to the aim of describing the evolution and development of cancer progression in terms of accumulation of genomic mutations over time (see an example in Figure \ref{fig:bn}).
In such a case, $V$ represents the set of genomic mutations, while $E$ represents the preferential ordering of accumulations among such mutations, depicted as relations of selective advantage in the graph \cite{loohuis2014inferring,ramazzotti2015capri,ramazzotti2016modeling}. 
% One example of cancer progression model is represented in Figure \ref{fig:bn}.
A progression model can also be encoded by using an adjacency matrix, which will be exploited here during the inference process, where the value 1 denotes that the genomic mutation of a given row favors the genomic mutation of the corresponding column.
Table \ref{tab:matrix} reports the binary adjacency matrix of the BN shown in Figure \ref{fig:bn}.

The input of the statistical methods used for the inference is an additional binary matrix $\mathbf{O} \in \{0,1\}^{M \times K}$ that contains the experimental observations.
This matrix is composed by $M$ rows, one for each observation (e.g., genomic data of a patient affected by cancer), and by
$K = |V|$ columns, which represent the genes whose progression has to be inferred. 
A value $o_{m,k} \in \mathbf{O}$ (with $m=1, \dots, M$ and $k=1, \dots, K$) is set to 1 when gene $k$ is mutated in the biological sample related to the $m$-th patient; on the contrary,  $o_{m,k} = 0$ to denote the absence of such mutation.

\begin{table}[!ht]
  \caption{Matrix representation of a BN.}
  \label{tab:matrix}
  \centering
\tiny
  \begin{tabular}{|c|c|c|c|c|c|c|c|c|c|c|c|c|c|c|c|}
  \hline
  &  \textbf{0} & \textbf{1} & \textbf{2} & \textbf{3} & \textbf{4} & \textbf{5} & 
  \textbf{6} & \textbf{7} & \textbf{8} & \textbf{9} & \textbf{10} & \textbf{11} & 
  \textbf{12} & \textbf{13} & \textbf{14} \\
  \hline
   \textbf{0} & 0 & \cellcolor{yellow}{1} &  0 & \cellcolor{yellow}{1} & 0 & 0 & 0 & \cellcolor{yellow}{1} & 0 & 0 & 0 & 0 & 0 & \cellcolor{yellow}{1} & \cellcolor{yellow}{1} \\
  \hline
   \textbf{1} & 0 & 0 &  0 & 0 & 0 & 0 & 0 & 0 & 0 & 0 & 0 & 0 & 0 &  0 & 0 \\
  \hline
   \textbf{2} & 0 & 0 &  0 & 0 & 0 & 0 & 0 & 0 & 0 & \cellcolor{yellow}{1} & 0 & 0 & 0 &  0 & 0 \\
  \hline
   \textbf{3} & 0 & 0 &  0 & 0 & 0 & 0 & 0 & 0 & 0 & 0 & 0 & 0 & 0 &  0 & 0 \\
  \hline
   \textbf{4} & 0 & 0 &  0 & 0 & 0 & 0 & 0 & 0 & \cellcolor{yellow}{1} & 0 & 0 & 0 & 0 &  0 & 0 \\
  \hline
   \textbf{5} & 0 & 0 &  0 & 0 & 0 & 0 & 0 & 0 & 0 & 0 & 0 & 0 & 0 &  0 & 0 \\
  \hline
   \textbf{6} & 0 & 0 &  0 & 0 & 0 & 0 & 0 & 0 & 0 & 0 & 0 & 0 & 0 &  0 & 0 \\
  \hline
   \textbf{7} & 0 & 0 &  \cellcolor{yellow}{1} & 0 & 0 & 0 & 0 & 0 & 0 & 0 & 0 & 0 & 0 &  0 & 0 \\
  \hline
   \textbf{8} & 0 & 0 &  0 & 0 & 0 & 0 & 0 & 0 & 0 & 0 & 0 & 0 & 0 &  0 & 0 \\
  \hline
   \textbf{9} & 0 & 0 &  0 & 0 & 0 & 0 & 0 & 0 & 0 & 0 & 0 & 0 & 0 &  0 & 0 \\
  \hline
   \textbf{10} & 0 & 0 &  0 & 0 & 0 & 0 & 0 & 0 & 0 & 0 & 0 & 0 & 0 &  0 & 0 \\
  \hline
   \textbf{11} & 0 & 0 &  0 & 0 & 0 & \cellcolor{yellow}{1}  & \cellcolor{yellow}{1}  & 0 & 0 & 0 & \cellcolor{yellow}{1} & 0  & \cellcolor{yellow}{1} &  0 & 0 \\
  \hline
   \textbf{12} & 0 & 0 &  0 & 0 & 0 & 0 & 0 & 0 & 0 & 0 & 0 & 0 & 0 &  0 & 0 \\
  \hline
   \textbf{13} & 0 & 0 &  0 & 0 & 0 & 0 & 0 & 0 & 0 & 0 & 0 & 0 & 0 &  0 & 0 \\
  \hline
\textbf{14} & 0 & 0 &  0 & 0 & \cellcolor{yellow}{1} & 0 & 0 & 0 & 0 & 0 & 0 & \cellcolor{yellow}{1} & 0 &  0 & 0 \\
  \hline

  \end{tabular}
\end{table}

Furthermore, state-of-the-art techniques that make use of BNs to model cancer progression, apply further constraints to the obtained network, leveraging the theory of probabilistic causation introduced by Patrick Suppes \cite{suppes1970probabilistic}. 
Specifically, Suppes defined the notion of \emph{prima facie causation}, where  a relation between a cause $u$ and its effect $v$ is verified when two conditions are observed to be true:
\begin{enumerate}
  \item  \emph{temporal priority} (TP): each cause must precede its effects;
  \item  \emph{probability raising} (PR): the presence of the causing event increases the probability of observing its subsequent effects.
\end{enumerate}

\noindent Given these conditions, we can formulate the following definition \cite{suppes1970probabilistic,ramazzotti2016modeling}:
\begin{definition}[Probabilistic causation~\cite{suppes1970probabilistic}] \label{def:praising}
For any couple of events $u$ and $v$, occurring respectively at times $t_u$ and $t_v$, under the mild assumptions that $0 < P(u), P(v) < 1$, the event $u$ is called a \emph{prima facie cause} of $v$ if it occurs \emph{before} $u$ and it \emph{raises the probability} of $u$:
\begin{equation}
\begin{cases}
(TP) \quad t_u < t_v \\
(PR) \quad P(v | u) > P(v | \neg u).
\end{cases}
\end{equation}
\end{definition}
% \todo[inline]{cos'e' $\overline{u}$}
Hence, the results of the statistical methods are constrained Bayesian networks---named Suppes-Bayes Causal Networks (SBCNs) \cite{ramazzotti2016modeling}---which account for the selective advantage relations among genomic events by combining probabilistic constraints
% (based on the theory of probabilistic causation by Patrick Suppes ) 
with maximum likelihood estimation \cite{suppes1970probabilistic}. 
Therefore, the problem of determining the sequence of mutations leading to cancer can be re-formulated as the problem of learning  the structure of a BN. 

There exist two strategies to tackle this structure inference problem: \emph{(i)} the constraint-based approaches, mainly due to the works of Judea Pearl, which consist in discovering the conditional independence relations within the input data to learn the BN \cite{Koller2009}; \emph{(ii)} the score-based approaches, in which the inference problem is re-stated as an optimization problem where all possible DAGs are considered valid solutions and they are evaluated using a likelihood-based score function \cite{Koller2009}. 

In this paper, we rely on the latter strategy. This approach assumes the data to be independent and identically distributed and, because of this, the likelihood of the data is the product of the likelihood of each datum, which, in turn, is defined by the factorized joint probability function described as follows:
\begin{eqnarray*}
P(\mathit{x_1, \ldots, x_k}) &=& \prod_{X_i \in V} P(X_i = x_i | Pa(X_i) = x_{Pa(i)}), \\
\end{eqnarray*}
where $x_1, \ldots, x_k$ are the nodes in the network and $Pa(.)$ indicates the parent set (i.e., all the nodes with an arch pointing to it) of a given node. 

For numerical reasons, log-likelihood ($LL$) is usually used instead of the likelihood itself, and thus the likelihood product becomes the log likelihood sum. To avoid overfitting, we use the so called Bayesian Information Criterion (BIC) \cite{bic_1978} ``regularized'' likelihood score function (which is calculated using the package \texttt{bnlearn} of the R programming tool \cite{Scutari2010}).
The score function is defined as follows: 
\begin{equation}\label{eq:bic}
score_{\text{BIC}}(D,G) = LL(D | G) - \dfrac{\log M}{2} \text{dim}(G),
\end{equation}
where $G$ denotes the considered DAG, $D$ denotes the input data, $M$ denotes the number of samples, and $\text{dim}(G)$ denotes the number of parameters of the DAG $G$.
Here, the parameters of $G$ refer to the set of arcs present in $G$ along with the encoded conditional probabilities.
It is worth noting that the regularization term $-\text{dim}(G)$ is exploited to favor nodes characterized by fewer parents, so that sparse graphs are promoted during the inference process.
This approach is adopted with the aim of providing models with the most confident set of arcs, even admitting the possibility of missing true relations. 
In fact, intuitively, we add to the inferred model only those arcs that strongly contribute to the calculation of the adopted likelihood score, which, in terms of likelihood, are also the most confident ones.

As the size of the sample employed in the inference process increases, both the weight of the regularization term and the weight of the likelihood function increase.
However, the increment of the latter is more relevant, so that with more input data the likelihood has a more pronounced contribution to the overall score function.
% With sample size enlarging, both the weight of the regularization term and the ``weight" of the likelihood increase with the latter increasing faster, hence, with more data, likelihood will contribute more to the score. 
Statistically speaking, we can state that the BIC is a consistent score \cite{Koller2009}, which means that having sufficiently large sample sizes, the network with the maximum BIC score is $I$-equivalent to the true generative structure \cite{Koller2009}. Specifically, two structures are said to be $I$-equivalent (where $I$ stands for independence), if they encode the exact same set of conditional independence relations among the considered variables. 
We remark that BNs with different structures may encode the same set of relations of dependency among variables. 
For this reason, BIC is said to converge to a solution I-equivalent to the ones generating the data, even if there is no guarantee to converge to the exact structure \cite{Koller2009}.

In general, independently from the strategy used in the inference process, the (huge) size of search space of valid solutions makes this problem very hard to tackle, especially from the computational time point of view.
% Regardless of the approach used for the inference, the huge search space of valid solutions is the main difficulty of this problem.

In particular, as stated above, the problem of BNs inference is NP-hard \cite{chickering2004large}; therefore all state-of-the-art techniques rely on heuristics  \cite{Koller2009}, and they are mainly based on stochastic population-based global optimization algorithms.
Specifically, methods based on Genetic Algorithms \cite{larranaga1996structure} and Ant Colony Optimization \cite{de2002ant,de2008learning,jun2009bayesian} have been proposed in the literature.
In this work we exploit the first methodology, described in the next section, and we investigate how the chosen heuristics impacts the performance. 

\subsection{Genetic Algorithms and Model Inference}
Genetic Algorithms (GAs) were introduced by J. H. Holland in 1975 \cite{holland75} as a global search methodology inspired by the mechanisms of natural selection.
In GA, a population $\mathcal{P}$ of candidate solutions iteratively evolves towards the global optimum of a user-specified fitness function.

GAs are characterized by a well-known convergence theorem named \textit{schema theorem}.
This theorem ensures that the presence of a schema (i.e., a template of solutions) in the population, having a good impact on the fitness value (i.e., the quality of a candidate solution),  increases exponentially generation after generation.
GAs were shown to be effective for the problem of the BN learning, both in the case of available and not available \emph{a priori} knowledge about nodes' ordering \cite{larranaga1996structure}, which allows a relevant reduction of the search space.
% This work is based on the previous experimentations of Larranaga \emph{et al}.

GAs are based on a population $\mathcal{P}$ composed of $Q$ randomly created individuals that are generally represented as fixed-length strings over a finite alphabet, encoding solutions of the problem under investigation.
In this work, each individual represents the linearized adjacency matrix of a candidate BN (e.g., the matrix in  Table \ref{tab:matrix}), by means of a string of binary values whose length is $K\times K$.
% This is the  characteristic that makes GAs particularly appealing for combinatorial optimization.

The individuals of the population undergo an iterative process whereby three genetic operators (selection, crossover, mutation) are applied, according to a given fitness function, to simulate the evolution process which results in a new population of possibly improved solutions.
The fitness function used in this work is the score formalized in Equation \ref{eq:bic}.
During the selection process, individuals from $\mathcal{P}$ are chosen and inserted into a new temporary population $\mathcal{P}^{'}$ using some fitness-dependent sampling procedure \cite{Baeck1994}.
In this work we assume a ranking selection: individuals are ranked according to their fitness values and the probability of selecting an individual is proportional to its \textit{position} in the ranking. 

The crossover operator is used to combine the structure of two promising parents into new and improved offspring, which are collected into a third population $\mathcal{P}^{''}$.
We assume a single point crossover, in which the two strings encoded by the two parents are ``cut'' in the same random position and one of the resulting substrings is exchanged.
The crossover is generally applied with a probability $p_c$; specifically, in our implementation, the crossover is  performed on the selected population with  $p_c = 1$, although it does not have an actual effect on the offspring when the cut position is equal to $0$ or equal to $K$.

Finally, the mutation operator is used to perturb the solutions encoded in the individuals of the population $\mathcal{P}^{''}$, thus allowing a further exploration of the search space.
Mutation alters a symbol of the individual, which is substituted with a random symbol from the alphabet with a fixed probability $p_m$.
In our tests mutation is applied by flipping a single bit of the individual with probability $p_m=0.01$, as suggested in \cite{larranaga1996structure}.

After the application of the genetic operators, GAs can proceed by following two alternative strategies: \emph{(i)} all individuals in $\mathcal{P}^{''}$ replace those in $\mathcal{P}$ (simple method); \emph{(ii)} the best $Q$ individuals in $\mathcal{P} \cup \mathcal{P}^{''}$ replace those in $\mathcal{P}$ (elitist method, which is exploited in this work).
% In this work, we rely on the elitist method.
Once the population is replaced, the process iterates until a halting  criterion is met, e.g., after a fixed number of generations (100 in this work).

It is worth noting that the one-point crossover and the mutation are not closed operators in the case of unordered nodes, since the resulting offspring might not encode valid DAGs.
To the aim of ensuring a consistent population of individuals throughout the generations, the two operators are followed by a correction procedure, in which we analyze the candidate BN to verify the potential  presence of cycles.
Our correction procedure works as follows: random arcs are removed from the solution until no more cycles are detected.
Cycles in the networks are detected using the \texttt{networkx} library \cite{hagberg-2008-exploring}; we exploit in particular the \verb!simple_cycles()! method, which returns a list of elementary circuits in the network identified using the Johnson's algorithm \cite{johnson1975finding}.
As long as the list is non-empty, we sample and remove arbitrary arcs. 
The obtained individual (i.e., a DAG) is finally added to the provisional population.
% \textbf{*QUINDI ALLA FINE L'INDIVIDUO POTATO E' QUELLO CHE RESTA NELLA POPOLAZIONE?*}

The number of fitness evaluations required by the overall evolutionary methodology---which is proportional to the number of generations and to the size of the population---can be very high, thus resulting computationally challenging.
% Due to the high number of fitness evaluations required (proportional to the number of generations and to the population size), the evolutionary approach can be computationally challenging. 
In order to mitigate this problem, our methodology was designed to exploit a high-performance architecture, described in the next section.

\subsection{Distributed Computing on GALILEO}
GALILEO is a Tier-1 supercomputer maintained by the Italian Consortium CINECA, devoted to scientific investigation.
This supercomputer is composed of 516 computing nodes; each computing node is equipped with two 8-core Haswell processors (clock frequency 2.4 GHz), for a total of 16 cores per computing node and an overall count of 8256 cores.
It is worth noting that hyperthreading is disabled on this computing nodes, reducing the overall level of parallelism to the physical 8 cores.
Computing nodes are also equipped with 128 GB of RAM (8 GB reserved for each core).
GALILEO also contains 768 Intel Xeon Phi 7120p co-processors and 20 GPUs Nvidia K80, although none of them was exploited in this work.

Our learning algorithm was implemented in a multi-threaded fashion, to fully leverage the cores of the computing nodes and distribute the fitness evaluations required by the GA.
As a further acceleration, we exploited the MPI library \cite{Dalcin2005} to distribute over several computing nodes the execution of multiple parallel optimizations, whose results are used to calculate the statistical data about the learning process.
% In particular, in this work we created \emph{in silico} $60$ synthetic causal trees characterized by 5 increasing levels of noise (for a total of $60 \times 5=300$ test cases), to investigate the performances of the learning algorithm.
As a last note, we point out that GALILEO's job scheduling system limits to 100 the number of simultaneous computing nodes that can be requested for a single job.
Hence, we subdivided all tests into separate jobs composed of 100 simultaneous optimizations.

\section{Results}\label{sec:results}
We tested the performance of our method both in terms of error rates between the inferred structure and the true one (i.e., the exact structure), and in terms of speed-up of the running time achieved with respect to a strictly sequential execution.
To this aim, we first generated a set of random structures, used as ground truth, representing BNs that are used as generative models for \emph{in silico} observations of genomic profiles.
Given these BNs, we sampled a set of datasets used as the starting point to perform the learning task. 
In order to mimic cancer progression and, specifically, its cumulative dynamics, we also constrained the conditional probabilities of the randomly generated BNs so that they only model positive dependencies among nodes.
Stated in other words, when generating the random structures to be learned, the random BNs model only situations where the presence of a parent node (positively) correlates (i.e., increases) the expected probability of later observing its child. 
So doing, we describe probabilistic relations of selective advantage among cancer clones, where the occurrence of an early mutation increases our expectation of observing, later on, its subsequent mutation during cancer progression \cite{ramazzotti2015capri}.

% In the following subsections, a detailed description of the results is provided. 

\subsection{Inference Results}
We first tested the performance of our method in terms of structural distance of the inferred solutions from the generative model (i.e., the exact structure). 
The performance is evaluated using the classic measures of accuracy, sensitivity and specificity defined as follows:
\begin{itemize}
  \item $accuracy = \frac{(TP + TN)}{(TP + TN + FP + FN)}$;
  \item $sensitivity = \frac{TP}{(TP + FN)}$;
  \item $specificity = \frac{TN}{(FP + TN)}$;
\end{itemize}
where $TP$ and $FP$ denote true and false positives, respectively, while $TN$ and $FN$ are the true and false negatives, respectively.
We define as positive any arc that is present in the generative model, while negative any arc that is not present in the generative model.
To be more precise, $TP$ are the arcs present in the generative model and correctly inferred by the method, while $FP$ are the arcs not inferred but present in the ground truth model.
On the contrary, $TN$ are the arcs not present in the generative model and (correctly) not inferred by the method, while $FN$ are the arcs not inferred but present in the ground truth model.
The measures of accuracy, sensitivity and specificity assume real values in $[0,1]$; a result $\approx 1$ indicates a good performance of our inference method. 

To assess the performance of our method we generated \emph{in silico} $200$ random trees composed of $15$ nodes (in this test we assign at most one predecessor per node).
Note that the size of random trees employed in our tests is analogous to the size of real models of cancer progression (see, for instance, \cite{10.1371/journal.pone.0027136}).
Moreover, to further complicate the problem, the $200$ random trees were subdivided in sets of $40$ synthetic causal trees characterized by 5 increasing levels of noise ($0\%, 5\%, 10\%, 15\%$ and $20\%$). %(for a total of $60 \times 5=300$ test cases).%, to investigate the performances of the learning algorithm.
Specifically, we applied a number of random perturbations according to the noise level to the observation matrices (i.e., bit flipping) to model any potential source of error that can occur during the experimental data collection. 
% In particular, we tested $5$ levels of noise: $0\%, 5\%, 10\%, 15\%$ and $20\%$.

Finally, for each tree, we generated a dataset of $M=100$ observations, encoded as binary matrices.
%, where the columns represent the mutated genes and the rows represent the patients. 
%In this matrix, an entry of value $1$ indicates that the relative mutation has been observed in the considered patient, while $0$ indicates the absence of such mutation as already defined in the problem definition. *QUESTO E' GIA' DETTO NELL'INTRO, PERO' IN TERMINI LEGGERMENTE DIVERSI. VA UNIFORMATO E SPIEGATO MEGLIO*}
We then executed the test of performances both for the reconstruction of constrained and unconstrained BNs (as described in \cite{ramazzotti2015capri}).
We want to point out that the methodology described in this paper is not limited to the inference of trees as it can, in principle, reconstruct any DAG \cite{loohuis2014inferring,larranaga1996structure}.
% To investigate the performances of our evolutionary methodology in presence of noise, we further complicated the problem by adding noise to the observations.
% Specifically, we applied a set of random perturbations in the observation matrices (i.e., bit flipping) to model any potential source of error during experimental data collection. 
% In particular, we tested $5$ levels of noise: $0\%, 5\%, 10\%, 15\%$ and $20\%$.
%, of random perturbations added to the observation matrices.

Tables \ref{tab:performance_distance_BN} and \ref{tab:performance_distance_SBCN} reports the results obtained for the inference of unconstrained BNs and SBCNs, respectively.
We observe high values of accuracy and specificity in both cases, which remain above $0.9$ in all conditions, demonstrating how our methodology is also robust to different levels of noise in the observed data.
In the case of SBCNs, however, we achieved better results, reaching a value of $0.97$ for the accuracy in the case of noise-free observations. 
SBCNs are also characterized by a value of $0.98$ for the specificity (i.e., capability of avoiding false positives) also assuming a relevant level of noise.
% \textbf{*NON C'E' UN REF CHE DICE CHE SOPRA 0.9 E' ACCURATO?*}. 
We observe that the values of sensitivity obtained in these tests are consistent with the expected performance of the adopted regularizator (i.e., BIC), which is specifically adopted here to converge to the most confident relations among genes. 
As a matter of fact, BIC is designed to produce sparse graphs, involving only the most likely arcs \cite{ramazzotti2016modeling}. For this reason, the sensitivity values (a measure of false negatives) are tipically lower than the ones of specificity (a measure of false positives), as BIC (at least with small sample sizes) tends to provide sparse networks that well describe the data, rather than predicting more arcs. If the goal was to perform prediction of possible unknown relations among genes while admitting false positives, other regularization such as the Akaike information criterion (AIC) \cite{akaike1998information}, may better fulfill this task.

\begin{table}[!htbp]
\centering
\caption{Accuracy, Sensitivity and Specificity of the GA applied to the problem of cancer progression inference tackled by means of Bayesian Networks.}
\label{tab:performance_distance_BN}
\begin{tabular}{ | c  | c |  c |  c |  c |  c  | }
    \hline
    & \multicolumn{5}{c|}{\bf{Noise}}\\	
	\hline
	 & $0\%$ & $5\%$ & $10\%$ & $15\%$ & $20\%$ \\ \hline
	 \hline
    \bf{Accuracy}  & $0.92$ & $0.92$ & $0.93$ & $0.92$ & $0.92$ \\ \hline
    \bf{Sensitivity} & $0.49$ & $0.49$ & $0.55$ & $0.51$ & $0.50$ \\ \hline
    \bf{Specificity} & $0.96$ & $0.96$ & $0.96$ & $0.96$ & $0.96$ \\ \hline
\end{tabular}
\end{table}

\begin{table}[!htbp]
\centering
\caption{Accuracy, Sensitivity and Specificity of the GA applied to the problem of cancer progression inference tackled by means of Suppes-Bayes Causal Networks.}
\label{tab:performance_distance_SBCN}
\begin{tabular}{ | c  | c |  c |  c |  c |  c  | }
    \hline
    & \multicolumn{5}{c|}{\bf{Noise}}\\	
	\hline
	 & $0\%$ & $5\%$ & $10\%$ & $15\%$ & $20\%$ \\ \hline
	 \hline
    \bf{Accuracy}  & $0.97$ & $0.97$ & $0.96$ & $0.95$ & $0.92$ \\ \hline
    \bf{Sensitivity} & $0.84$ & $0.83$ & $0.76$ & $0.71$ & $0.58$ \\ \hline
    \bf{Specificity} & $0.98$ & $0.98$ & $0.98$ & $0.98$ & $0.95$ \\ \hline
\end{tabular}
\end{table}

%\todo[inline]{TODO: attendere risultati con priori, aggiornare tabella, commentare le differenze. Inoltre verificare bene quante simulazioni sono inserite nella versione finale del paper: nel testo è segnato 300 random trees.}

\subsection{Computational Performances}
GAs belong to the class of iterative and population-based optimization meta-heuristics.
Thus, during each generation, the fitness evaluations must be calculated for each individual.
Since all individuals are mutually independent, the process of fitness evaluations can be parallelized.

\begin{figure}[!htbp]
\centering
\includegraphics[width=.49\textwidth]{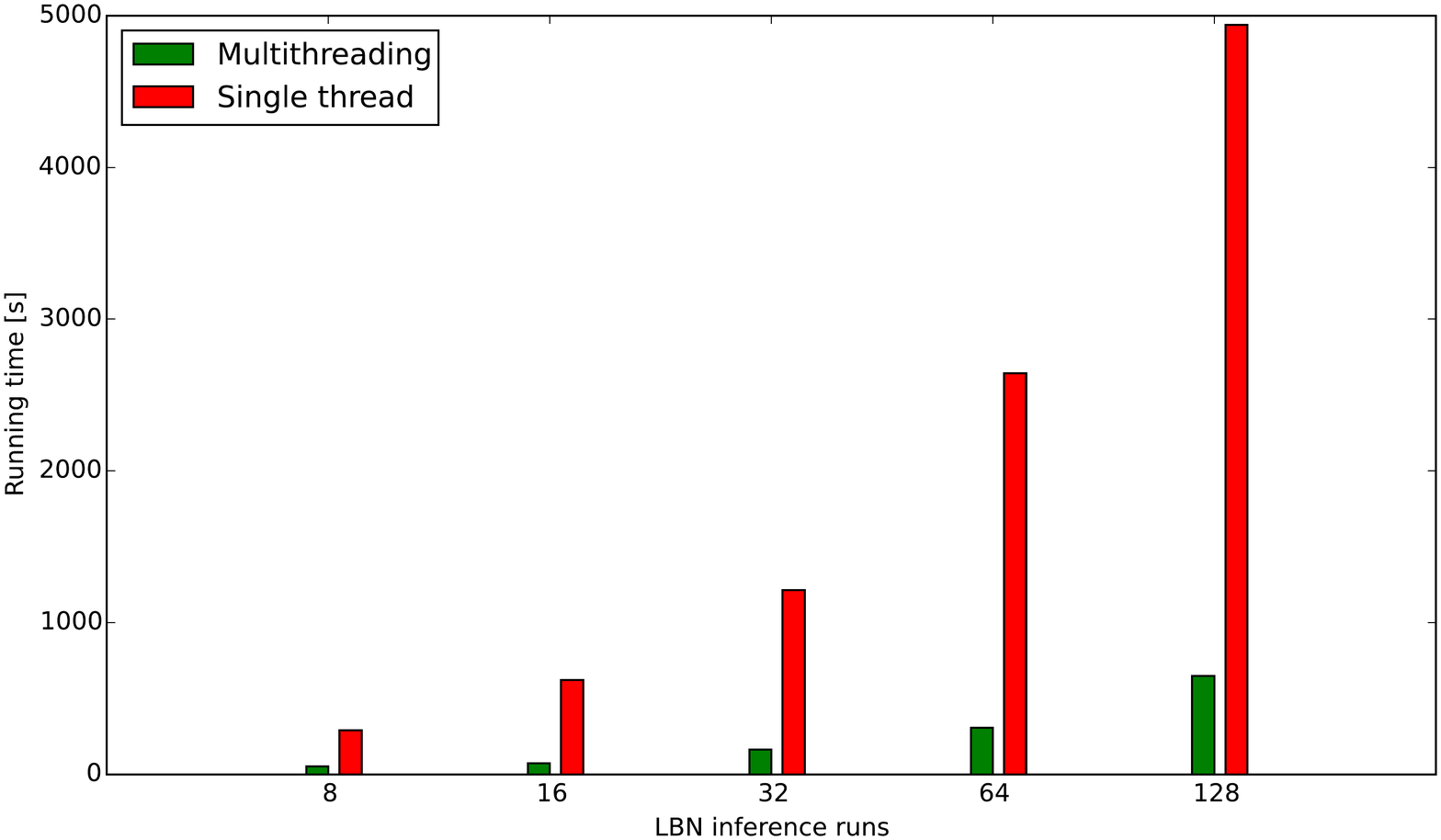}
\caption{Comparison of the running time (in seconds) of the learning algorithm using distributed fitness evaluations  (yellow bars) with respect to a classic single threaded execution (red bars).}
\label{fig:multithread}
\end{figure}

In this work, the fitness evaluations---which are based on BIC---are performed using the package \texttt{bnlearn} of the R programming tool \cite{Scutari2010}. 
Multiple threads are created, each one running an instance of \texttt{bnlearn}, to independently calculate the fitness values. 
A join mechanism allows to synchronize the termination of all threads, collect the results and update the fitness values for all individuals at once.
By using this multi-threaded strategy we obtained a speed-up---with respect to a strictly sequential execution on the same computing node---approximatively equal to  $8\times$ with a populations $\mathcal{P}$ of the GA with a number of individuals $Q > 8$ (see Figure \ref{fig:multithread}). 
For instance, in the case of the test with $Q = 128$ individuals, the running time of the optimization was reduced from 1 hours and 22 minutes down to 647 seconds, corresponding to a $7.6\times$ speed-up. 

\begin{figure}[!htbp]
\centering
\includegraphics[width=.49\textwidth]{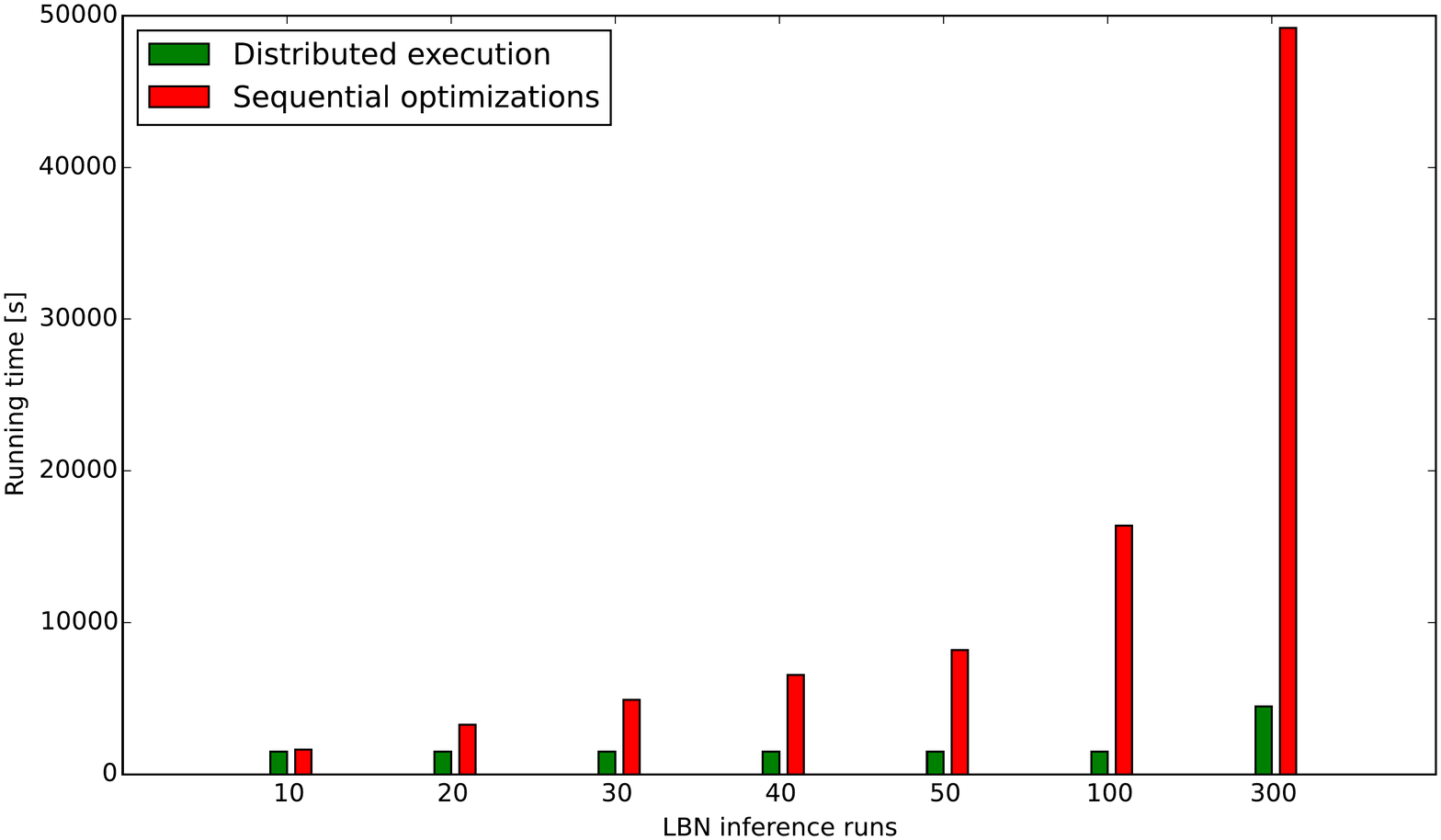}
\caption{Comparison of the running time (in seconds) of the learning algorithm distributed on the GALILEO supercomputer (yellow bars) with respect to a simple multi-threaded execution (red bars).}
\label{fig:multinode}
\end{figure}

The second level of parallelism was introduced by means of  MPI, which was used to distribute the GA instances over multiple computing nodes.
The results in Figure \ref{fig:multinode} summarize the speed-up obtained using this approach, highlighting that for a few optimizations the overhead due to MPI reduces the potential acceleration. 
However, when the number of parallel optimizations is greater than 10, the distributed architecture  strongly reduces the execution time of the inference processes. 
Not surprisingly, the maximum acceleration is achieved in the case of 100 simultaneous GA executions, completed in 24 minutes, where the same job executed on a single node takes 4 hours and a half, corresponding to a $11 \times$ speed-up. 
Clearly, the restriction on the number of computing nodes that can be concurrently employed on GALILEO supercomputer limits the performance gain that can be achieved.
Indeed, in the case of $200$ optimizations we still achieved an overall speed-up equal to $11 \times$ (see Figure \ref{fig:multinode}).

% \todo[inline]{aggiornare i tempi in funzione delle 200 ottimizzazioni}
% Because of GALILEO's jobs limitations, in the case of 300 optimizations we still achieved an overall speedup equal to $11 \times$. 

In conclusion, the relevant reduction of the running time provided by supercomputers can be exploited to assess the confidence in the estimation of an inferred network. 
As a matter of fact, as discussed in details in \cite{caravagna2015algorithmic}, the statistic support of a learned network can be calculated by performing multiple optimizations and by repeatedly sampling the input data (namely, bootstrap or cross-validation approaches). 
In this case, it is suggested to perform at least $100$ repetitions of the optimization process, which would benefit from our distributed approach.

\section{Conclusion}\label{sec:concl}
In this work we presented a methodology for the efficient inference of models of cancer progression from genomic data, and we assessed its performances in terms of accuracy, sensitivity and specificity, showing the good results related to the inferred BNs and the robustness of the optimization process in the case of noisy experimental data.

The methodology exploited in this paper is based on Genetic Algorithms \cite{larranaga1996structure}, and it is accelerated by means of the combination of multi-threading and distributed computation. 
Thanks to our approach, the overall computation time was reduced of almost two orders of magnitude using a parallel architecture (CINECA's GALILEO): on the one hand, the multi-threaded execution of the fitness functions allowed  a $7.6 \times$  speed-up on each computing node; on the other hand, the parallel execution of multiple optimizations distributed over   independent nodes allowed a further $11 \times$ speed-up, for an overall reduction of the execution time of approximately $84 \times$.
% \textbf{*QUINDI QUESTO E' IL DATO RISPETTO AL SINGLE THREAD? SE SI, VA DETTO ANCHE NEI RESULTS*}. 

It is worth noting that GALILEO is equipped with additional parallel co-processors, namely GPUs and MICs.
Both architectures are characterized by a theoretical peak power of about one  tera-flop.
By further distributing the calculations on these co-processors, we could strongly reduce the execution time. 
In particular, GPUs typically contain thousands computing cores that can be used to calculate in a parallel fashion the fitness functions required by the GA, allowing to deal with larger populations of individuals whose optimization would require a reduced running time with respect to the parallel strategy employed in this paper (we refer the interested reader to \cite{Nobile2016} for a review of GPU-powered methods for computational biology). 
Thus, as future development of this work, we plan to port the overall methodology to the CUDA architecture. 

In our optimization method, both mutation and crossover are non closed operators in the case of non-ordered nodes. 
This means that they might introduce cycles in the BN, which are identified using Johnson's algorithm and corrected by means of random arcs removal.
The complexity of the cycle finding algorithm is $\mathcal{O}((v+e)(c+1))$, where $v=|V|$, $e=|E|$ and $c$ is the number of elementary circuits in the graph.
However, all these computations are not necessary, because we just need to know that \emph{at least} one cycle exists in the BN. 
We will therefore modify the cycle finding algorithm to improve the efficiency of the correction step exploited in this work.

% As a final remark, in our representation, the values on the diagonal belong to the candidate solution even though they must remain equal to zero.
% However, mutation can introduce a direct arc from a node to itself: this affects both the explorative capabilities of the operator and the performances of the algorithm. 


\section*{Acknowledgment}
The authors would like to thank A. Tangherloni for his suggestions about MPI development.
We acknowledge the CINECA award under the ISCRA initiative, for the availability of high performance computing resources and support.



\bibliographystyle{IEEEtran}
\bibliography{cibcb-lbn}
\end{document}